\documentclass[final]{cvpr}
\usepackage{times}
\usepackage{epsfig}
\usepackage{graphicx}
\usepackage{amsmath}
\usepackage{amssymb}

\usepackage{verbatimbox}
\usepackage{multirow}
\usepackage{tabu}
\usepackage[ruled,vlined]{algorithm2e}
\usepackage{algpseudocode}
\usepackage[pagebackref=false,breaklinks=true,colorlinks,bookmarks=false]{hyperref}

\def\HW(#1){\textcolor{red}{HW: #1}}
\def\MAC(#1){\textcolor{magenta}{MAC: #1}}
\def\TJ(#1){\textcolor{blue}{TJ: #1}}
\def\YF(#1){\textcolor{green}{YF: #1}}

\def\cvprPaperID{2236} % *** Enter the CVPR Paper ID here
\def\confYear{CVPR 2021}
\newcommand{\forcond}{$k=1$ \KwTo $K$}

\begin{document}

%%%%%%%%% TITLE
\title{BASAR:Black-box Attack on Skeletal Action Recognition}

\author{Yunfeng Diao$^{1,2}$\thanks{The research was conducted during the visit to the University of Leeds.}, \ \ \ Tianjia Shao$^3$\thanks{Corresponding author}, \ \ \ Yong-Liang Yang$^4$, \ \ \ Kun Zhou$^3$, \ \ \ He Wang$^1$\thanks{https://youtu.be/PjWgwnAkV8g}\\
$^1$University of Leeds, UK\ \ \ \ $^2$Southwest Jiaotong University, China\\ $^3$State Key Lab of CAD\&CG, Zhejiang University, China\ \ \ \ $^4$ University of Bath, UK\\
{\small dyf@my.swjtu.edu.cn, tjshao@zju.edu.cn, y.yang@cs.bath.ac.uk, kunzhou@zju.edu.cn, h.e.wang@leeds.ac.uk}

% \and
}

\maketitle
%no page number for camera ready
\pagestyle{empty}
\thispagestyle{empty}
%%%%%%%%% ABSTRACT
\begin{abstract}
   Skeletal motion plays a vital role in human activity recognition as either an independent data source or a complement~\cite{minh_dang_sensor-based_2020}. The robustness of skeleton-based activity recognizers has been questioned recently \cite{liu_adversarial_2019,wang_smart_2020}, which shows that they are vulnerable to adversarial attacks when the full-knowledge of the recognizer is accessible to the attacker. However, this white-box requirement is overly restrictive in most scenarios and the attack is not truly threatening. In this paper, we show that such threats do exist under black-box settings too. To this end, we propose the first black-box adversarial attack method BASAR. Through BASAR, we show that adversarial attack is not only truly a threat but also can be extremely deceitful, because on-manifold adversarial samples are rather common in skeletal motions, in contrast to the common belief that adversarial samples only exist off-manifold \cite{gilmer_adversarial_2018}. Through exhaustive evaluation and comparison, we show that BASAR can deliver successful attacks across models, data, and attack modes. Through harsh perceptual studies, we show that it achieves effective yet imperceptible attacks. By analyzing the attack on different activity recognizers, BASAR helps identify the potential causes of their vulnerability and provides insights on what classifiers are likely to be more robust against attack. Code is available at \href{https://github.com/realcrane/BASAR-Black-box-Attack-on-Skeletal-Action-Recognition}{https://github.com/realcrane/BASAR-Black-box-Attack-on-Skeletal-Action-Recognition}.
   
   %Human Activity Recognition (HAR) serves as an essential component in safety, security, autonomous vehicles, etc, in which skeletal motions play a vital role as either an independent data source or a way to improve the robustness \cite{minh_dang_sensor-based_2020}. Recently, it has been shown that skeleton-based action recognizers are vulnerable to adversarial attack \cite{liu_adversarial_2019,wang_smart_2020}. However, existing attack methods are essentially white-box approaches and require full-knowledge of the attacked model. This makes it hard to apply them in real-world scenarios, raising the question of whether it is a real threat. In this paper, we show that such threats do exist. We propose the first black-box adversarial attack method, BASAR, on skeleton-based action recognizers. We show that attacking skeletal motions is significantly different from attacking other data, leading to a new line of research. We also demonstrate that on-manifold adversarial samples are rather common in skeletal motions, in contrast to the common belief that adversarial samples are off-manifold \cite{gilmer_adversarial_2018}. Through exhaustive evaluation and comparison, we show BASAR's superior performance across models, data and attack modes. By harsh perceptual studies, we show that BASAR can achieve effective yet imperceptible attacks.
\end{abstract}

%%%%%%%%% BODY TEXT
\section{Introduction}

Deep learning methods have been proven to be vulnerable to carefully devised data perturbations since first identified in \cite{szegedy_intriguing_2014}. This causes major concerns especially in safety and security \cite{akhtar_survey_2018}, as the perturbations are \textit{imperceptible} to humans but \textit{destructive} to machine intelligence.
Consequently, how to detect and defend attacks has also been investigated \cite{chakraborty_adversarial_2018}. While the attack on static data (e.g. images, texts, graphs) has been widely studied, the attack on time-series data has only been recently explored \cite{karim_adversarial_2020,fawaz_adversarial_2019}. In this paper, we look into a specific yet important type of time series data, skeletal motions, under adversarial attack.

Skeletal motion is crucial in activity recognition as it increases the robustness by mitigating issues such as lighting, occlusion, view angles, etc \cite{ren_survey_2020}. Therefore, the vulnerability of skeleton-based classifiers under adversarial attack has recently drawn attention \cite{liu_adversarial_2019,zheng_towards_2020,wang_smart_2020}. Albeit identifying a key issue that needs to be addressed, their methods are essentially \textit{white-box} methods. The attempt on black-box attack is via surrogate models, i.e. attack a classifier in a white-box manner then use the results to attack the target classifier. While white-box attack requires the full knowledge of the attacked model which is unlikely to be available in real-world scenarios, black-box attack via surrogate models cannot guarantee success due to its heavy dependence on the choice of the surrogate model \cite{wang_understanding_2021}. In this paper, we propose BASAR, the \textit{very first} black-box attack method on skeletal action recognition to our best knowledge.

A skeletal motion has unique features that distinguish itself from other data under adversarial attack. First, a skeleton usually has less than 100 Degrees of freedom (Dofs), much smaller than previously attacked data such as images/meshes. This low dimensionality leads to low-redundancy \cite{tramer_space_2017}, restricting possible attacks within small subspaces. Second, \textit{imperceptibility} is a prerequisite for any successful attack, but its evaluation on skeletal motions is under-explored. Different from the attack where visual imperceptibility has high correlations mainly with the perturbation magnitude (e.g. images), a skeletal motion has dynamics that are well-recognized by human perceptual systems. More specifically, any sparse attack, e.g. on individual joints or individual frames, albeit small would break the dynamics and therefore be easily perceptible. In contrast, coordinated attacks on all joints and frames can provide better imperceptibility even when perturbations are relatively large \cite{wang_smart_2020}. As a result, using the perturbation magnitude alone (as in most existing methods) is not a reliable metric for skeletal motion. Last but not least, prior methods mainly assume that adversarial samples are off the data manifold \cite{gilmer_adversarial_2018}. As we will show, skeletal motion is one real-world example where on-manifold adversarial samples not only exist but are rather common. This raises a serious concern for human activity recognition solutions as these on-manifold adversarial samples are \textit{implementable}.

Given a motion $\mathbf{x}$ with class label $C_\mathbf{x}$, BASAR aims to find $\mathbf{x}'$ that is close to $\mathbf{x}$ (measured by some distance function) and can fool the black-boxed classifier such that $C_{\mathbf{x}'}\ne C_\mathbf{x}$. BASAR formulates it as a constrained optimization problem, aiming to find $\mathbf{x}'$ that is just outside $C_\mathbf{x}$ while still on the data manifold. The optimization is highly non-linear due to the complexity of the classification boundaries and the data manifold. The former dictates that any greedy search (e.g. gradient-based) near the boundaries will suffer from local minima; while the latter means that not all perturbation directions result in equal visual quality (in-manifold perturbation tends to be better than off-manifold perturbation). Consequently, there are often conflicts between these two spaces when searching for $\mathbf{x}'$. To reconcile the conflicts, we propose a method called \textit{guided manifold walk} (GMW). GMW consists of three sub-routines: aimed probing, random exploration, and manifold projection. It starts from a random position (untargeted attack) outside $C_\mathbf{x}$, or a random sample within a specific class (targeted attack with the specific class as the targeted class). It then can approach $\mathbf{x}$ by aimed probing attempting to find a sample which is close to the boundary of $C_\mathbf{x}$, or by random explorations to overcome local minima to find samples that are closer to $\mathbf{x}$, or by manifold projection to find the closest point on the data manifold. The above sub-routines are iteratively executed driven by the quality of the adversarial sample until a satisfactory $\mathbf{x}'$ is found or the maximum number of iterations is reached.

We extensively evaluate BASAR on several state-of-the-art methods using multiple datasets in both untargeted and targeted attack tasks. The results show that not only is BASAR successful across models and datasets, it can also find on-manifold adversarial samples, in contrast to the common assumption that adversarial samples only exist off-manifold \cite{gilmer_adversarial_2018}. On par with very recent work that also found on-manifold samples in images \cite{stutz_Disentangling_2019}, we show, for the first time, the existence and commonality of such samples in skeletal motions. We also comprehensively compare BASAR with other methods, showing the superiority of BASAR by large margins. Finally, since the perturbation magnitude alone is not enough to evaluate the attack quality, we conduct harsh perceptual studies to evaluate the naturalness, deceitfulness, and indistinguishability of the attack.

Formally, we demonstrate that adversarial attack is truly a threat to skeleton-based activity recognition. To this end, we propose the first black-box attack method and comprehensively evaluate the vulnerability of several state-of-the-art activity recognition methods. We show the existence of on-manifold adversarial samples in various skeletal motion datasets and provide key insights on what classifiers tend to resist on-manifold adversarial samples.

\section{Related Work}
\subsection{Skeleton-based Activity Recognition}

%Early research in this area has well explored a variety of hand-crafted features~\cite{Vemulapalli:2014:HAR,Fernando:2015:MVE,Devanne:2015:HAR} to recognize human actions from skeleton data. Benefiting from trained features of deep neural networks, recent  deep learning based methods have achieved the state-of-the-art performance. According to the actual skeleton data representation for learning, these methods can be categorized into sequence-based, image-based, and graph-based methods, respectively.

%Sequence-based methods treat the skeleton data as a chronological sequence of skeleton joint coordinates. The RNN architectures can in turn be employed to perform the classification~\cite{Du:2015:HRN,Liu:2016:STL,Song:2017:ESA,Zhang:2019:VAN}. Image-based methods encode the skeleton motion using a 2D pseudo-image, where one dimension represents time, and the other dimension stacks all joints of a single skeleton. As such, CNN-based image classification can be adopted~\cite{Liu:2017:ESV,Ke:2017:NRS,Kim_2017_CVPR_Workshops}. Graph-based methods consider the skeleton as a topological graph where the nodes correspond to joints and edges correspond to the bones connecting neighboring joints. Then the graph convolutional networks (GCNs) are used to effectively recognize the actions~\cite{Yan:2018:STG,Li_2019_CVPR,Shi_2019_CVPR,Shi:2019:SBA,Cheng_2020_CVPR,Zhang_2020_CVPR_2}. %

Early research has well explored a variety of hand-crafted features~\cite{Vemulapalli:2014:HAR,Fernando:2015:MVE,Devanne:2015:HAR} to recognize human actions from skeletal motions. Benefiting from the ability of neural networks, recent deep learning based methods have achieved state-of-the-art performance. Among them, RNN-based methods treat the motion data as a time series of joint coordinates, and employ common RNN architectures such as LSTMs to conduct the motion classification~\cite{Du:2015:HRN,Liu:2016:STL,Song:2017:ESA,Zhang:2019:VAN}. CNN-based methods try to utilize image classification methods to recognize the motion data~\cite{Liu:2017:ESV,Ke:2017:NRS,Kim_2017_CVPR_Workshops}. To this end, they transform the motion data into 2D pseudo-images, where each column stores a frame of skeleton joint coordinates. Graph convolutional network (GCN) based methods consider the skeleton as a topological graph where the nodes correspond to joints and edges correspond to the bones connecting neighboring joints. They perform convolution operations on graphs to effectively recognize the actions~\cite{Yan:2018:STG,Li_2019_CVPR,Shi_2019_CVPR,Shi:2019:SBA,Cheng_2020_CVPR,Zhang_2020_CVPR_2}. 
For example, to capture long-range joint relations, Liu et al.~\cite{Liu_2020_CVPR} utilize disentangled multi-scale graph convolutions, which contains a unified spatial-temporal GCN operator for capturing complex spatial-temporal dependencies. Zhang et al.~\cite{Zhang_2020_CVPR_1} introduce joint semantics to the GCN model, resulting in a lightweight yet effective method. 

Our work is complementary to existing research by exploring their vulnerability to adversarial attacks and suggesting potential improvements. Based on the codes shared by the authors, we extensively evaluated BASAR on several state-of-the-art methods, demonstrating that even the latest methods with remarkable successes are still vulnerable to adversarial attacks.

\subsection{Adversarial Attack}
\subsubsection{White-box Attack}

A large variety of adversarial attack methods have been explored in different tasks~\cite{akhtar_survey_2018,sun2018adversarial,2020Adversarial}. Most of them consider the white-box setting, where the full-knowledge of the classifier is known by the attacker. Apart from common vision-based tasks such as classification~\cite{goodfellow_explainingah_2015,kurakin_advphy_2017,carlini_towards_2017}, adversarial attacks on time series and action recognition have attracted attention recently. For general time series, multivariate time-series attack has been conducted~\cite{liu_adversarial_2019, harford2020adversarial} based on adversarial transformation networks (ATN) \cite{baluja2017adversarial}. Untargeted attack is proposed for optical flow based action recognition \cite{inkawhich_adversarial_2018}. For skeletal action recognition, a one-step and an iterative method are proposed \cite{liu_adversarial_2019} based on FGSM and PGD~\cite{kurakin_advphy_2017} respectively to attack GNN-based models. Wang et al. \cite{wang_smart_2020} achieve better attacking results using a novel perceptual loss that minimizes the dynamics difference between the original and the adversarial motion. Moreover, the performance is proven by extensive perceptual studies. Zheng et al. \cite{zheng_towards_2020} formulate adversarial attack as an optimization problem with joint angle constraints in the Euclidean space, then solve it with Alternating Direction Method of Multipliers.

Although white-box attacks achieve satisfactory performance based on the full access to the attacked models, they are not very applicable in real-world scenarios since the details of classifiers are not usually exposed to the attacker. Hence, it is unclear whether the classifier vulnerability under (white-box) adversarial attack is a real threat. We on the other hand show that black-box attack is possible and is more threatening.

\subsubsection{Black-box Attack}
The difficulty of applying white-box attack in the real-world motivates the study of black-box attack, in which the attackers cannot access any underlying information of the attacked model but only make queries. A simple approach is transfer-based attack, which generates adversarial samples from one surrogate model via white-box attack~\cite{papernot_transfer_2016,papernot_bbattack_2017}. Existing black-box methods on skeletal motion~\cite{liu_adversarial_2019, wang_smart_2020,zheng_towards_2020} all rely on such a method, and cannot guarantee success due to the heavy dependence on the surrogate model~\cite{wang_smart_2020}. Another type of methods is to use the predicted scores or soft labels (e.g. the softmax layer output) for attack \cite{chen2017zoo, tu_autozoom_2019}. However, this is not exactly a black-box setting since the attacker can still access the model.

In a truly black-box setting, only the final class labels (hard-labels) can be used. Brendel et al. \cite{drendel_decisionaa_2018} perform the first hard-label attack by a random walk along the decision boundary. Optimization-based approaches select the search direction by estimating the gradient using the information computed at the decision boundary~\cite{chen_hopskipjump_2020, cheng_signopt_2020, DBLP:conf/iclr/ChengLCZYH19}. Dong et al.~\cite{Dong2019EfficientDB} model the local geometry of the search directions through an evolutionary attack algorithm. Li et al.~\cite{Li2020QEBAQB} explore the subspace optimization methods from three different representative subspaces including spatial, frequency, and intrinsic component.
%In contrast, Chen et al. \cite{chen_hopskipjump_2020} select the search direction by estimating the gradient using the information computed at the decision boundary. Cheng et al. \cite{cheng_signopt_2020} formulate the hard-label attack as an optimization problem by applying a zeroth-order optimization algorithm.

Hard-label black-box attack on time-series data is much less explored. In the black-box setting in~\cite{harford2020adversarial, karim_adversarial_2020}, to replicate/approximate the target model, the authors train a local substitute student network to mimic the final decision of the teacher classifier, to attack general time series. To attack videos, Wei et al. \cite{wei_heuristicbb_2020} heuristically attack the key frames and regions based on \cite{cheng_signopt_2020} due to the high dimensionality. To our best knowledge, there is no prior work on hard-label black-box attack on skeletal activity recognition.

\section{Methodology}
We denote a motion with $n$ frames as $\mathbf{x}$ = $\{x^1,\dots, x^n\}$, each frame $x^t$ = $\{q_1,\dots,q_m\}$ including $m$ Dofs. These Dofs are usually joint positions or angles, depending on the data. Existing action recognizers take motions as input and output a class label. Specifically, given a trained recognizer $G$, $G$ maps a motion to a probabilistic distribution over classes, $G$: $\mathbf{x}\rightarrow R^k$ where $k$ is the total number of action classes. The class label $C_\mathbf{x}$ then can be derived e.g. via \textit{softmax}. To find an adversarial sample $\mathbf{x'}$ of motion $\mathbf{x}$, we start with the formulation in \cite{carlini_towards_2017} but augment it for modeling motions:
\begin{align}
     & \text{minimize }\ \ L(\mathbf{x}, \mathbf{x'})  \nonumber \\
     & \text{subject to }\ \ C_{\mathbf{x'}} = c \text{ and } \mathbf{x'} \in [0, 1]^{m\times n},
\end{align}
where $L$ is the Euclidean distance. $c$ is the targeted class. Note that the constraint can also be replaced by $C_{\mathbf{x'}} \ne C_{\mathbf{x}}$ for untargeted attack. However, simply applying this formulation to skeletal motions is not sufficient because it only restricts the adversarial sample $\mathbf{x'}$ in a hyper-cube $[0,1]^{m\times n}$. Given that human poses lie in a natural pose manifold $\mathcal{M}$ \cite{wang_STRNN_2019, wang_harmonic_2013}, $\mathbf{x'}$ can easily contain off-manifold frames which are unnatural/implausible poses and easily perceptible. We therefore add another constraint $\mathbf{x'} \in \mathcal{M}$:
\begin{align}
\label{eq:opt}
     & \text{minimize }\ \ L(\mathbf{x}, \mathbf{x'}) \nonumber \\ & \text{subject to }\ \   \mathbf{x'} \in [0, 1]^{m\times n}, \mathbf{x'} \in \mathcal{M} \nonumber \\
    & C_{\mathbf{x'}} = c\text{ (targeted)} \text{ or } C_{\mathbf{x'}} \ne C_{\mathbf{x}}\text{ (untargeted)}. 
\end{align}
In practice, we find that $\mathbf{x'} \in [0, 1]^{m\times n}$ is less restrictive than other constraints and always satisfied. Eq.~\ref{eq:opt} is highly nonlinear and cannot be solved analytically. It thus requires a numerical solution. 

\begin{figure}[t]
    \centering
    \includegraphics[width=0.8\linewidth]{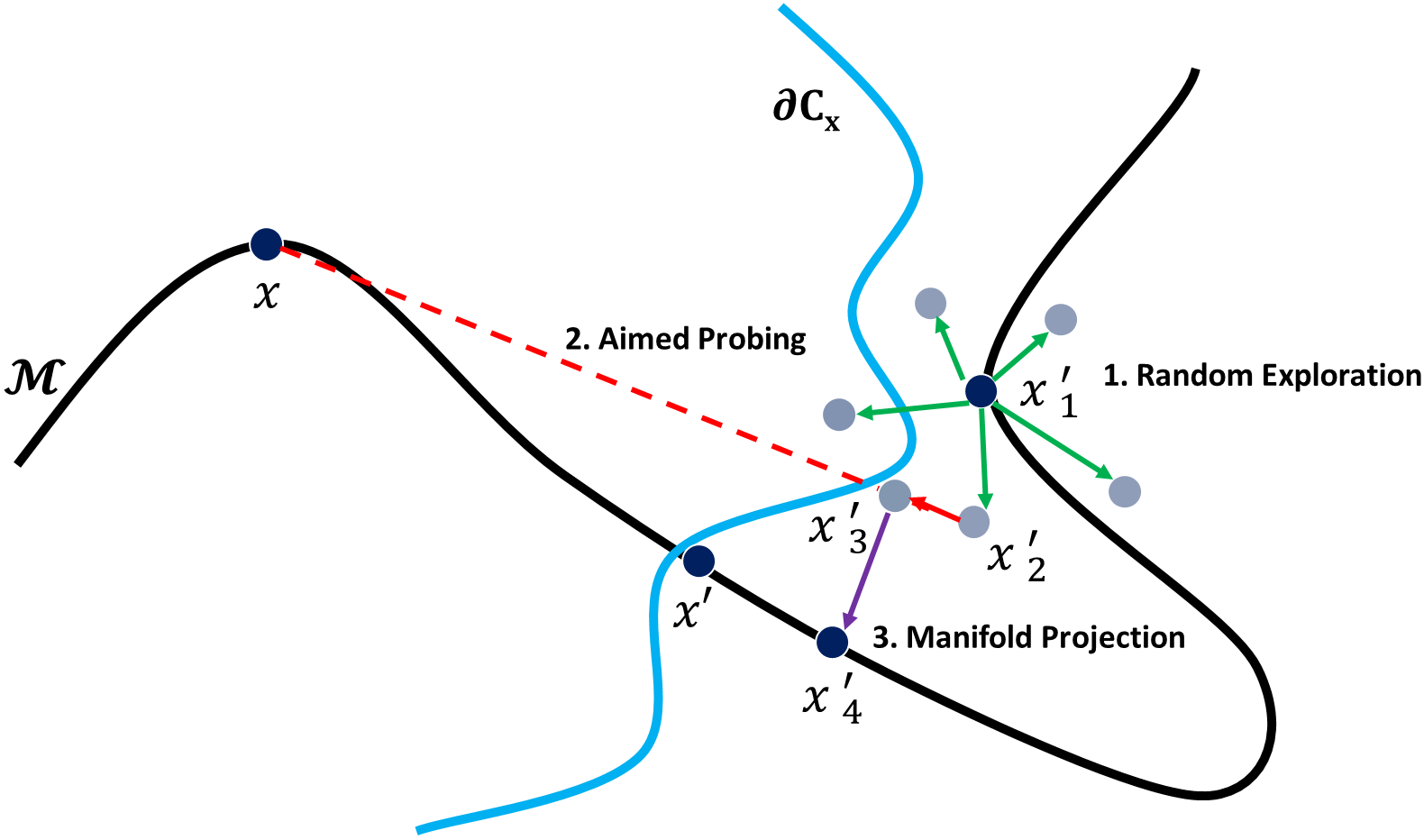}
    \caption{2D illustration of BASAR on a single frame. $x$ is the attacked frame. $x'$ is the ideal adversarial sample. $\mathcal{M}$ (black line) is the natural pose manifold and $\partial C_{\mathbf{x}}$ (blue line) is the class boundary of $C_{\mathbf{x}}$. $x'_1$ is the result of last iteration. $x'_2$, $x'_3$ and $x'_4$ are the three intermediate results of the current iteration.}
    \label{fig:overall}
    \vspace{-0.3cm}
\end{figure}

\subsection{Guided Manifold Walk}
We propose a new method called \textit{Guided Manifold Walk} (GMW) to solve Eq.~\ref{eq:opt}. For simplicity, we start with a single frame $x$ and a 2D  illustration of GMW on $x$ shown in Fig.~\ref{fig:overall}, where $x'$ is the ideal adversarial sample which is on-manifold and close to $x$.  Given the non-linearity of the classification boundary and the data manifold, BASAR aims to exploit the properties of both simultaneously, inspired by decision-boundary-based attack \cite{drendel_decisionaa_2018}. In each iteration, BASAR conducts three sub-routines: \textit{random exploration}, \textit{aimed probing}, and \textit{manifold projection}. Random exploration is to explore the vicinity of the current adversarial sample $x'_1$ to find a random sample $x'_2$. Aimed probing is to find a sample $x'_3$ in proximity to $\partial C_{\mathbf{x}}$ and is closer to $x$, in the direction from $x'_2$ to $x$. Finally, manifold projection is to project $x'_3$ onto $\mathcal{M}$ to obtain $x'_4$. Although GMW is only explained for a single frame $x$, it can be applied to a motion $\mathbf{x}$. The algorithm overview for perturbing the whole motion is given in Alg.~\ref{al1}, where $\lambda$ and $\beta$ are parameters and $l$ is a distance function. They are defined in Eq.~\ref{eq:re},~\ref{eq:ap} and ~\ref{eq:metrics} respectively. Next, we give details of random exploration, aimed probing, and manifold projection. 

% %algorithm 1
% \begin{algorithm}[t]
% \SetAlgoLined
% \label{al1}
% \textbf{Input}: attacked motion $\mathbf{x}$; random sample $\widetilde{\mathbf{x}}_0$, $C_{\widetilde{\mathbf{x}}_0} = c\text{ (targeted)}
%  \text{ or } C_{\widetilde{\mathbf{x}}_0} \ne C_{\mathbf{x}}\text{ (untargeted)}$; maximum number of iterations $K$; threshold $\epsilon$\;
% \textbf{Initialization}: $\mathbf{x}'_0 = Aimed Probing (\widetilde{\mathbf{x}}_0 , \mathbf{x})$, ensuring that $\mathbf{x}'_0$ is adversarial\;
%  \For{\forcond}{
%     \Repeat{$\widetilde{\mathbf{x}}$ is adversarial \text{ or } $\lambda < 10^{-10}$ }{
%         $\widetilde{\mathbf{x}} = Random Exploration (\mathbf{x}'_{k-1}, \mathbf{x})$\;
%                 }
%     \leIf{$\lambda \ge 10^{-10}$}{$\mathbf{x}'_k$ = $\widetilde{\mathbf{x}}$}{break}
%     \Repeat{$\widetilde{\mathbf{x}}$ is adversarial \text{ or } $\beta < 10^{-10}$ }{
%         $\widetilde{\mathbf{x}} = Aimed Probing (\mathbf{x}'_{k}, \mathbf{x})$\;
%                 }
%     \leIf{$\beta \ge 10^{-10}$}{$\mathbf{x}'_k$ = $\widetilde{\mathbf{x}}$}{break}
%     ${\mathbf{\hat{x}}} = Manifold Projection (\mathbf{x}'_{k}, \mathbf{x})$\;
%     \While{$\mathbf{\hat{x}}$ is not adversarial}{
%       $\mathbf{\hat{x}} = Aimed Probing (\mathbf{x}'_{k}, \mathbf{\hat{x}})$\;
%     }
%     \leIf{l($\mathbf{\hat{x}}$,$\mathbf{x}$) $\ge \epsilon$}{$\mathbf{x}'_k$ = $\mathbf{\hat{x}}$}{break}
% }
%  \caption{Overview of the GMW}
% \end{algorithm}

\subsection{Random Exploration} 
We first extend the image-based random exploration in \cite{drendel_decisionaa_2018} to motions with dynamic weighting. Random exploration is an operation to explore in proximity to the classification boundary, by making a small step towards a random direction. The random perturbation is calculated by Eq.~\ref{eq:re}: 
\begin{align}
\label{eq:re}
& \widetilde{\mathbf{x}} =\mathbf{x}' + \mathbf{W}\Delta, \nonumber \\
& \text{where }\Delta _{*} = \mathbf{R}_{*} - (\mathbf{R}_{*}^T\mathbf{d}_{*})\mathbf{d}_{*}, \  \mathbf{d}_{*} = \frac{\mathbf{x}_{*} - \mathbf{x}'_{*}}{\| \mathbf{x}_{*} - \mathbf{x}'_{*}\|}, \nonumber \\
& \mathbf{R}_{*} = \lambda \frac{\mathbf{r}}{\| \mathbf{r}\|} \| \mathbf{x}_{*} - \mathbf{x}'_{*}\|\text{, } r \in N(0, \mathbf{I}),
\end{align}
% \begin{equation}
% \label{eq:re}
% \widetilde{\mathbf{x}} =\mathbf{x}' + \mathbf{W}\Delta, 
% \end{equation}
% $ \text{where }\Delta _{*} = \mathbf{R}_{*} - (\mathbf{R}_{*}^T\mathbf{d}_{*})\mathbf{d}_{*}, \  \mathbf{d}_{*} = \frac{\mathbf{x}_{*} - \mathbf{x}'_{*}}{\| \mathbf{x}_{*} - \mathbf{x}'_{*}\|}, \nonumber \\
%  \mathbf{R}_{*} = \lambda \frac{\mathbf{r}}{\| \mathbf{r}\|} \| \mathbf{x}_{*} - \mathbf{x}'_{*}\|\text{, } r \in N(0, \mathbf{I}),$
where $\widetilde{\mathbf{x}}$ is the new perturbed sample, $\mathbf{x}$ and $\mathbf{x}'$ are the attacked motion and current adversarial sample. We use joint positions and the subscript $*$ indicates either the $x$, $y$, or $z$ joint coordinate. Unlike \cite{drendel_decisionaa_2018}, the update on $\mathbf{x}'$ is $\Delta$ weighted by $\mathbf{W}$ - a diagonal matrix with joint weights. This is because we observe that perturbations on different joints are not equally effective and imperceptible, e.g. perturbation on the spinal joints cause larger visual distortion but is less effective. We therefore weight them differently. $\Delta_\textit{*}$ controls the direction and magnitude of the update, and depends on two variables $\mathbf{R}_*$ and $\mathbf{d}_*$. $\mathbf{d}_*$ is the directional vector from $\mathbf{x}'$ to $\mathbf{x}$. $\mathbf{R}_*$ is a random directional vector sampled from a Normal distribution $N(0, \mathbf{I})$ where $\mathbf{I}$ is an identity matrix, $\mathbf{I} \in R^{z \times z}$, $z = mn/3$, $m$ is the number of Dofs in one frame and $n$ is total frame number. This directional vector is scaled by $\| \mathbf{x}_{*} - \mathbf{x}'_{*}\|$ and $\lambda$.

In Eq.~\ref{eq:re}, $\Delta$ is \textit{orthogonal} to $\mathbf{d}$. The random exploration essentially explores in the directions that are orthogonal to the direction towards $\mathbf{x}$, within the local area around $\mathbf{x}'$: $\mathbf{x}' + \mathbf{W} \Delta$. This mainly aims to avoid local minima along the $\partial C_{\mathbf{x}}$ when approaching $\mathbf{x}$. If $\widetilde{\mathbf{x}}$ is not adversarial, we decrease $\lambda$ and perform the random exploration again; otherwise, we increase $\lambda$, then enter the next sub-routine.
%algorithm 1
\begin{algorithm}[t]
\SetAlgoLined
\label{al1}
\textbf{Input}: attacked motion $\mathbf{x}$; random sample $\widetilde{\mathbf{x}}_0$, $C_{\widetilde{\mathbf{x}}_0} = c\text{ (targeted)}
 \text{ or } C_{\widetilde{\mathbf{x}}_0} \ne C_{\mathbf{x}}\text{ (untargeted)}$; maximum number of iterations $K$; threshold $\epsilon$\;
\textbf{Initialization}: $\mathbf{x}'_0 = Aimed Probing (\widetilde{\mathbf{x}}_0 , \mathbf{x})$, ensuring that $\mathbf{x}'_0$ is adversarial\;
 \For{\forcond}{
 
    \Repeat{$\widetilde{\mathbf{x}}$ is adversarial \text{ or } $\lambda < 10^{-10}$ }{
        $\widetilde{\mathbf{x}} = Random Exploration (\mathbf{x}'_{k-1}, \mathbf{x})$\;
                }
    \leIf{$\lambda \ge 10^{-10}$}{$\mathbf{x}'_k$ = $\widetilde{\mathbf{x}}$}{break}
    \Repeat{$\widetilde{\mathbf{x}}$ is adversarial \text{ or } $\beta < 10^{-10}$ }{
        $\widetilde{\mathbf{x}} = Aimed Probing (\mathbf{x}'_{k}, \mathbf{x})$\;
                }
    \leIf{$\beta \ge 10^{-10}$}{$\mathbf{x}'_k$ = $\widetilde{\mathbf{x}}$}{break}
    ${\mathbf{\hat{x}}} = Manifold Projection (\mathbf{x}'_{k}, \mathbf{x})$\;
    \While{$\mathbf{\hat{x}}$ is not adversarial}{
      $\mathbf{\hat{x}} = Aimed Probing (\mathbf{x}'_{k}, \mathbf{\hat{x}})$\;
    }
    \leIf{l($\mathbf{\hat{x}}$,$\mathbf{x}$) $\ge \epsilon$}{$\mathbf{x}'_k$ = $\mathbf{\hat{x}}$}{break}
}
 \caption{Overview of the GMW}
\end{algorithm}

\subsection{Aimed Probing}
We adopt Aimed probing from \cite{drendel_decisionaa_2018}, aiming to find a new adversarial sample between the perturbed motion and the original, so that the new adversarial sample is closer to the attacked motion:
\begin{equation}
\label{eq:ap}
\widetilde{\mathbf{x}} = \mathbf{x}' + \beta(\mathbf{x} - \mathbf{x}'),
\end{equation}
where $\beta$ is a forward step size. $\beta$ is decreased to conduct the aimed probing again if $\widetilde{\mathbf{x}}$ is not adversarial; otherwise, we increase $\beta$, then enter the next sub-routine. %{\color{red} Different from \cite{drendel_decisionaa_2018}, we dynamically control $\beta$ which is decreased to conduct the aimed probing again if $\widetilde{\mathbf{x}}$ is not adversarial; otherwise, we increase $\beta$, then enter the next sub-routine.}

\subsection{Manifold Projection}
After aimed probing and random exploration, the perturbed motion $\widetilde{\mathbf{x}}$ is often off the manifold, resulting in implausible and unnatural poses. We thus project them back to the manifold. Natural pose manifold can be obtained in two ways: explicit modeling \cite{wang_energy_2015} or implicit learning \cite{wang_STRNN_2019,Chen_dynamic_20}. Using implicit learning would require to train a data-driven model then use it for projection \cite{wang_STRNN_2019}, breaking BASAR into a two-step system. Therefore we employ explicit modeling. Specifically, we replace the constraint $\mathbf{x}' \in \mathcal{M}$ in Eq.~\ref{eq:opt} with hard constraints on bone lengths and joint rotations. We also constrain the dynamics of $\mathbf{x}'$ to be similar to $\mathbf{x}$:
\begin{align}
\label{eq:gmw_Euclidean}
    & \min_\mathbf{x'}\ \ L(\widetilde{\mathbf{x}}, \mathbf{x'})  + w L(\mathbf{\ddot x}, \mathbf{\ddot x'})  \nonumber \\ 
    & \text{subject to}\ \ B'_{i} = B_i \text{ and } \theta_i^{\min} \le \theta'_i \le \theta_i^{\max} \nonumber \\
    & C_{\mathbf{x'}} = c\text{ (targeted)} \text{ or } C_{\mathbf{x'}} \ne C_{\mathbf{x}}\text{ (untargeted)},
\end{align}
where $\mathbf{\ddot x}$ and $\mathbf{\ddot x'}$ are the $2nd$-order derivatives of $\mathbf{x}$ and $\mathbf{x}'$, $w$ is a weight. Matching the $2nd$-order derivatives is proven to be important for visual imperceptibility in adversarial attack \cite{wang_smart_2020}. $L$ is the Euclidean distance. $B_i$ and $B'_i$ are the $i$-th bone's lengths of the attacked and adversarial motion respectively. When the bone lengths change from frame to frame in the original data, we impose the bone-length constraint on each frame. $\theta'_i$ is the $i$-th joint in every frame of $\mathbf{x'}$ and subject to joint limits bounded by $\theta_i^{\min}$ and $\theta_i^{\max}$. 

Eq.~\ref{eq:gmw_Euclidean} is difficult to solve, especially to satisfy both the bone length and joint limit constraints in the joint position space \cite{zheng_towards_2020}. We therefore solve Eq.~\ref{eq:gmw_Euclidean} in two steps. First, we solve it without any constraints by Inverse Kinematics \cite{wang_energy_2015} in the joint angle space, which automatically preserves the bone lengths. Next, Eq.~\ref{eq:gmw_Euclidean} is solved in the joint angle space:
\begin{align}
\label{eq:gmw_angles}
    & \min_{\mathbf{\theta'}}\ \ L(\widetilde{\mathbf{\theta}}, \mathbf{\theta'})  + w L(\mathbf{\ddot \theta}, \mathbf{\ddot \theta'})  \nonumber \\ 
    & \text{subject to}\ \ \theta_i^{\min} \le \mathbf{\theta'_i} \le \theta_i^{\max}, \nonumber \\
    & C_{\mathbf{x'}} = c\text{ (targeted)} \text{ or } C_{\mathbf{x'}} \ne C_{\mathbf{x}}\text{ (untargeted)}.
\end{align}
Note that the objective function in Eq.~\ref{eq:gmw_angles} is designed to match the joint angles and the joint angular acceleration. We use a primal-dual interior-point method \cite{hedengren2014nonlinear,wachter2006implementation} to solve Eq.~\ref{eq:gmw_angles}. After solving for $\theta'$, the joint positions of the adversarial motion are computed using Forward Kinematics. Please refer to the supplementary materials for the details of mathematical deduction, implementation, and performance.

% \subsection{Implement Details}
% To adjust $\lambda$, we execute $q$ times random explorations instead of one time in a sub-routine call, and compute the attack successful rate of the $q$ intermediate samples. If the probability is less than 40\%, we reduce $\lambda$; if it is higher than 60\% we increase it; otherwise we do not update $\lambda$. For targeted attack, we randomly select one from all adversarial samples to do aimed probing. For untargeted attack, the $q$ adversarial samples may output $p$ labels. Therefore we randomly select one in every type of adversarial labels. After doing aimed probing to the $p$ adversarial samples with  different labels respectively, we only keep the one that has the smallest distance to original motion $x$. Also, when the adversarial sample is close to the original motion, we set a threshold value $\epsilon$ to make sure $\lambda$ is not higher than $\epsilon$. This is to ensure that the attack can eventually converge. In the end, considering the computing consumption, it is unrealistic to execute manifold projection in every iteration. We usually execute manifold projection after multiple iterations of aimed probing and random exploration.

\section{Experiments}
\subsection{Target Models and Datasets}
We select three state-of-the-art action recognition models to attack, STGCN~\cite{Yan:2018:STG}, MS-G3D~\cite{Liu_2020_CVPR} and SGN~\cite{Zhang_2020_CVPR_1}. STGCN~\cite{Yan:2018:STG} is one of the first GCN-based methods. MS-G3D~\cite{Liu_2020_CVPR} and SGN~\cite{Zhang_2020_CVPR_1} are both state-of-the-art methods. The three methods are sufficiently representative. We choose three frequently used benchmark datasets: HDM05~\cite{cg-2007-2}, NTU60~\cite{shahroudy2016ntu} and Kinetics~\cite{kay2017kinetics}. HDM05 dataset~\cite{cg-2007-2} has 130 action classes, 2337 sequences from 5 non-professional actors. Due to its high quality, any obvious difference between the adversarial motion and original motion can be easily perceived, making HDM05 suitable for our perceptual study. Since our target models are not trained on HDM05 in the original papers, we process HDM05 following~\cite{Du:2015:HRN} and train the target models strictly following the protocols in the papers, achieving 87.2\%, 94.4\%, 94.1\%  on STGCN, MS-G3D and SGN respectively. NTU RGB+D 60~\cite{shahroudy2016ntu} includes 56568 skeleton sequences with 60 action classes, performed by 40 subjects. Due to the large intra-class and viewpoint variations, it is ideal for verifying the effectiveness and generalizability of our approach. Kinetics 400~\cite{kay2017kinetics} is a large but highly noisy human action video dataset taken from different YouTube Videos. The skeletons are extracted from Openpose~\cite{cao2017realtime}, consisting of over 260000 skeleton sequences. Since SGN is not trained on it, we follow \cite{Zhang_2020_CVPR_1} to preprocess Kinetics skeletons and then train the model. Because some target models are big and making queries to them become slow, it is impractical to attack all motions in a dataset. So we randomly sample motions to attack. We gradually increase the number of motions to attack until all evaluation metrics (explained below) stabilize, so that we know the attacked motions are sufficiently representative in the dataset. 

% \YF (The SGN model was not trained on it, therefore we follow \cite{Zhang_2020_CVPR_1} to preprocess Kinetics Skeleton and then train the model, which achieve 37.5\% classification accuracy.) 

\subsection{Evaluation Metrics}
We employ the success rate as a major indicator for evaluation. In addition, to further numerically evaluate the quality of the adversarial samples, we also define evaluation metrics between the original motion $\mathbf{x}$ and its adversarial sample $\mathbf{x}'$, including the averaged joint position deviation $l$, the averaged joint acceleration deviation $\Delta a$, the averaged joint angular acceleration deviation $\Delta\alpha$, and the averaged bone-length deviation percentage $\Delta B/B$:
\begin{align}
\label{eq:metrics}
l =  \frac{1}{n N} \sum_{j=0}^{N} \| \mathbf{x}^{(j)} - \mathbf{x}^{'(j)}\|_2   \nonumber \\ 
\Delta a = \frac{1}{n O N} \sum_{j=0}^{N}\| \mathbf{\ddot x}^{(j)} - \mathbf{\ddot x}^{'(j)}\|_2  \nonumber \\
\Delta \alpha =\frac{1}{n O N} \sum_{j=0}^{N}\| \mathbf{\ddot \theta}^{(j)} - \mathbf{\ddot \theta}^{'(j)}\|_2  \nonumber\\  
\Delta B/B = \frac{\sum_{j=0}^{N}\sum_{i=0}^{T}((B_{i}^{(j)}-B_{i}^{'(j)}) / B_i^{(j)})}{TN} 
\end{align}
where $N$ is the number of adversarial samples. $O$ and $T$ are the total number of joints and bones in a skeleton. $n$ is the number of frames in a motion. In addition, we also investigate the percentage of on-manifold (OM) adversarial motions after the attack. We regard a motion as on-manifold if all its frames respect bone-length and joint limit constraints. Finally, since Kinetics has missing joints from time to time, it is impossible to attack it in the joint angle space. So we only attack it in the joint position space. Consequently, $\Delta\alpha$ and OM cannot be computed on Kinetics.

\subsection{Untargeted Attack}
\begin{table}
\centering
\setlength{\tabcolsep}{1mm}{
\begin{tabular}{|c|c|ccccc|}
\hline
Models&  & $l$$\downarrow$ & $\Delta a$$\downarrow$ & $\Delta\alpha\downarrow$ & $\Delta B/B$$\downarrow$ & OM$\uparrow$ \\
\hline
\multirow{2}{*}{STGCN} & MP & 0.13  & 0.05 & 0.11 & 0.00\% & 99.55\% \\
& No MP & 0.10 & 0.04 & 0.34 & 0.66\% & 0.00\%\\
\hline
\multirow{2}{*}{MSG3D} & MP & 0.76  & 0.12 & 0.49  & 1.78\% & 0.13\% \\
& No MP & 0.70  & 0.09 & 0.82  & 1.81\% & 0.00\%\\
\hline
\multirow{2}{*}{SGN} & MP & 11.53 & 1.92 & 6.70 & 9.60\% & 60.52\% \\
& No MP & 7.93 & 2.00 & 14.36 & 39.64\% & 0.00\%  \\ 
\hline\hline
\multirow{2}{*}{STGCN} & MP & 0.08 & 0.02 & 0.07 & 4.82\% & 4.68\%  \\
& No MP & 0.10 & 0.02 & 0.09 & 5.57\% & 1.82\%  \\ 
\hline
\multirow{2}{*}{MSG3D} & MP & 0.08 & 0.03 & 0.12 & 8.14\% & 0.86\%  \\
& No MP & 0.12 & 0.03 & 0.17 & 10.02\% & 0.57\%  \\ 
\hline
\multirow{2}{*}{SGN} & MP & 0.28 & 0.08 & 0.21 & 11.11\% & 28.95\% \\
& No MP & 0.30 & 0.10 & 0.42 & 28.00\% & 4.55\%  \\ 
\hline\hline
\multirow{2}{*}{STGCN} & MP & 0.05 & 0.0057 & n/a & 2.54\% & n/a  \\
& No MP & 0.07 & 0.0062 & n/a & 3.53\% & n/a   \\ 
\hline
\multirow{2}{*}{MSG3D} & MP & 0.10 & 0.011 & n/a & 5.16\% & n/a  \\
& No MP & 0.10 & 0.012 & n/a & 5.69\% & n/a   \\ 
\hline
\multirow{2}{*}{SGN} & MP & 0.12 & 0.020 & n/a & 4.23\% & n/a  \\
& No MP & 0.13 & 0.022 & n/a & 6.93\% & n/a   \\ 
\hline
\end{tabular}}
\caption{Untargeted attack on HDM05 (top), NTU (middle) and Kinetics (bottom). All attacks have a 100\% success rate. $l$: averaged joint position deviation; $\Delta$a: averaged joint acceleration deviation; $\Delta\alpha$: averaged joint angular acceleration deviation; $\Delta$B/B: averaged bone-length deviation percentage; on-manifold sample percentage (OM). MP means Manifold Projection.}
\label{tab:untargeted}
\vspace{-0.3cm}
\end{table}

To initialize for untargeted attack, we randomly sample a motion $\mathbf{x}'$ for a target motion $\mathbf{x}$ where $C_{\mathbf{x}'} \ne C_\mathbf{x}$. For HDM05, we randomly select 700 motions to attack STGCN, MSG3D and SGN. For NTU and Kinetics, we randomly sample 1200 and 500 motions respectively. The maximum iterations are 500, 1000 and 2000 on HDM05, Kinetics and NTU respectively. 

The results are shown in Tab.~\ref{tab:untargeted}. Note that BASAR achieves 100\% success in all tasks. Here we also conduct ablation studies (MP/No MP) to show the effects of manifold projection. First, the universal successes across all datasets and models demonstrate the effectiveness of BASAR. The manifold projection directly affects the OM results. BASAR can generate as high as 99.55\% on-manifold adversarial samples. As shown in the perceptual study later, the on-manifold samples are very hard to be distinguished from the original motions even under very harsh visual comparisons.  Detailed confusion matrices, visual results and ablation studies can be found in the supplementary materials. We leave the performance variation analysis across models to Section \ref{sec:robustness}.

\subsection{Targeted Attack}
\begin{table}
\centering
\setlength{\tabcolsep}{1mm}{
\begin{tabular}{|c|c|ccccc|}
\hline
Models&  & $l$$\downarrow$ & $\Delta a$$\downarrow$ & $\Delta\alpha\downarrow$ & $\Delta B/B$$\downarrow$ & OM$\uparrow$ \\
\hline
\multirow{2}{*}{STGCN} & MP & 4.97  & 0.10 & 0.65  & 3.44\% & 56.98\% \\
& No MP & 6.25  & 0.09 & 0.92  & 5.85\% & 0.00\%  \\ 
\hline
\multirow{2}{*}{MSG3D} & MP & 4.34  & 0.12 & 0.71  & 4.51\% & 1.64\%  \\
& No MP & 4.35  & 0.11 & 1.01  & 5.08\% & 0.00\%  \\ 
\hline
\multirow{2}{*}{SGN} & MP & 16.31 & 1.28 & 6.97 & 12.29\% & 20.96\% \\
& No MP& 16.13 & 1.63 & 13.28 & 29.86\% & 0.00\%  \\ 
\hline\hline
\multirow{2}{*}{STGCN} & MP & 0.37 & 0.03 & 0.25 & 9.73\% & 0.63\%  \\
& No MP & 0.38 & 0.04 & 0.16 & 11.55\% & 0.16\%  \\ 
\hline
\multirow{2}{*}{MSG3D} & MP & 0.36 & 0.05 & 0.24 & 15.43\% & 0.00\%  \\
& No OP & 0.40 & 0.06 & 0.27 & 17.72\% & 0.00\%  \\ 
\hline
\multirow{2}{*}{SGN} & MP & 1.28 & 0.09 & 0.38 & 28.24\% & 2.63\%  \\
& No MP & 1.35 & 0.10 & 0.53 & 39.43\% & 0.00\%  \\
\hline\hline
\multirow{2}{*}{STGCN} & MP & 0.63 & 0.03 & n/a & 29.10\% & n/a  \\
& No MP & 0.67 & 0.03 & n/a & 31.48\% & n/a   \\ 
\hline
\multirow{2}{*}{MSG3D} & MP & 0.56 & 0.05 & n/a & 27.26\% & n/a  \\
& No MP & 0.57 & 0.07 & n/a & 28.35\% & n/a   \\ 
\hline
\multirow{2}{*}{SGN} & MP & 1.51 & 0.18 & n/a & 68.45\% & n/a  \\
& No MP & 1.54 & 0.19 & n/a & 72.09\% & n/a   \\ 
\hline
\end{tabular}}
\caption{Targeted attack on HDM05 (top), NTU (middle) and Kinetics (bottom). All attacks have a 100\% success rate.}
\label{tab:targeted}
\vspace{-0.3cm}
\end{table}

For targeted attack, we randomly select the same number of motions from each dataset as in untargeted attack, with maximum iteration set to 3000. To initiate a targeted attack on motion $\mathbf{x}$, we randomly select a motion $\mathbf{x}'$ where $C_{\mathbf{x}'}$ = $c$ and $c$ is the targeted class.

The results are shown in Tab.~\ref{tab:targeted}. All attacks achieve 100\% success. The targeted attack is more challenging than the untargeted attack \cite{wang_smart_2020}, because the randomly selected label often has completely different semantic meanings from the original one. Attacking an `eating' motion to `drinking' is much easier than to `running'. This is why the targeted attack in general has worse results than untargeted attack under every metric. Even under such harsh settings, BASAR can still produce as high as 56.98\% on-manifold adversarial samples. The performance variation across models is consistent with the untargeted attack. Additional details can be found in the supplementary materials.

\subsection{Classifier Robustness}
\label{sec:robustness}
We notice that BASAR's performance varies across target models consistently in both the untargeted and targeted attack. In other words, action recognizers are not equally gullible. SGN, in general, is the hardest to fool across all datasets. When looking at the results, the perturbation always needs to be larger to fool SGN, shown by larger $l$, $\Delta$a, $\Delta\alpha$ and $\Delta$B/B on every dataset compared with STGCN and MSG3D. This includes both joint-angle and joint-position attack. We speculate that it has to do with the features that SGN uses. Unlike STGCN and MSG3D which use raw joints and bones and rely on networks to learn good classification features, SGN also employs \textit{semantic} features, where different joint types are encoded in learning their patterns and correlations. This type of semantic information requires large perturbations to bring the motion out of its pattern. Hence, semantic information improves its robustness against attacks, as larger perturbations are more likely to be perceptible. Note that SGN sometimes has a higher OM percentage. However, we find that these OM motions have large visual differences from the original motions. In other words, unlike STGCN and MSG3D where a higher OM percentage indicates more visually indistinguishable adversarial samples, the OM samples of SGN look natural, can fool the classifier and probably can fool humans when being observed independently, but are unlikely to survive strict side-by-side comparisons with the original motions. Next, MSG3D is slightly harder to fool than STGCN. Although both use joint positions, MSG3D explicitly uses the bone information, which essentially recognizes the relative movements of joints. The relative movement pattern of joints helps resist attack by requiring the perturbation to be large enough to break the patterns. Finally, the robustness of a specific model varies across datasets. Meanwhile, the pattern of robustness variations across datasets also differs from model to model. It is hard to theoretically identify the cause and we will leave it to future research. Additional detailed findings can be found in the supplementary materials.

\subsection{Perceptual Study}
Numerical evaluation alone is not sufficient to evaluate the imperceptibility of adversarial attack on skeletal motions, because they cannot accurately indicate whether the attack is perceptible to humans \cite{wang_smart_2020}. We, therefore, conduct three perceptual studies: Deceitfulness, Naturalness, Indistinguishability, following \cite{wang_smart_2020}. Since we have 36 scenarios (models vs datasets vs attack types vs MP/No MP), it is unrealistic to exhaustively cover all conditions. We choose HDM05 and untargeted attack for our perceptual studies. We exclude NTU and Kinetics as they contain severe and noticeable noises. Our preliminary study shows that it is hard for people to tell if a motion is attacked therein. The quality of HDM05 is high  where perturbations can be easily identified. In total, we recruited 50 subjects (age between 20 and 54). We briefly explain our study setting here and refer the readers to the supplementary materials for details.

We first test whether BASAR visually changes the meaning of the motion and whether the meaning of the original motion is clear to the subjects (\textbf{Deceitfulness}). In each user study, we randomly choose 45 motions (15 from STGCN, MSG3D and SGN respectively) with the ground-truth label and after-attack label for 45 trials. In each trial, the video is played for 6 seconds then the user is asked to choose which label best describes the motion with no time limits. We also perform ablation studies to test whether on-manifold adversarial samples look more natural than off-manifold samples (\textbf{Naturalness}). In each trial, we randomly select 120 motions for 60 trials. Each trial includes one motion attacked with manifold projection and one without. The two motions are played side-by-side for 6 seconds twice, then the user needs to pick which one looks more natural, or cannot tell the difference, with no time limits. Finally, we conduct the strictest test to see whether the adversarial samples from BASAR can survive side-by-side scrutiny (\textbf{Indistinguishability}). In each trial, two motions are displayed side by side. The left motion is always the original and the user is told so. The right one can be the original motion or the attacked motion. The user is asked whether he/she can see any visual difference after the motions are played for 6 seconds twice. 

\subsubsection{Results}
\begin{figure}[t]
\centering
\includegraphics[width=0.8\linewidth]{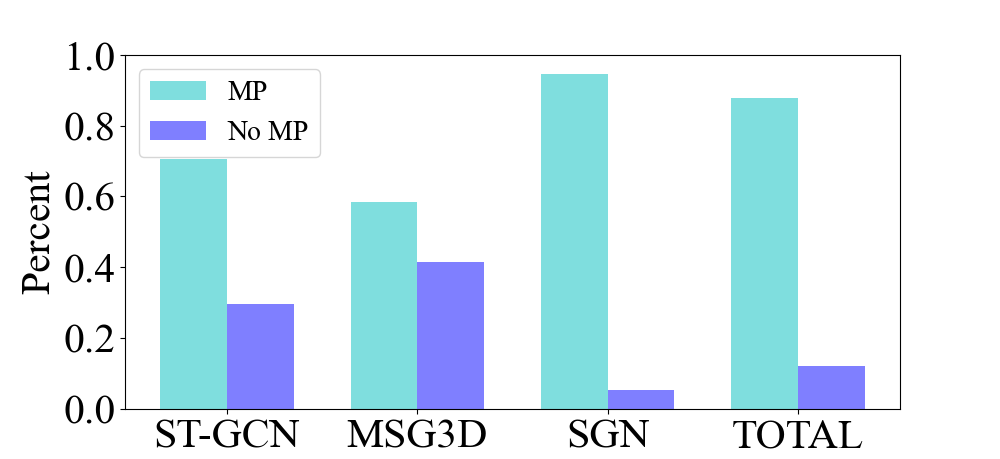}
   \caption{Normalized user preferences on Naturalness. MP/No MP refers to with/without manifold projection. The vertical axis is the percentage of user preference. The horizontal axis is the attacked models.}
\label{fig:naturalness}
\vspace{-0.5cm}
\end{figure}

The average success rate of \textbf{Deceitfulness} is 79.64\% across three models, with 88.13\% , 83.33\%, 67.47\% on STGCN, MS-G3D and SGN respectively. This is consistent with our prediction because SGN requires larger perturbations, thus is more likely to lead to the change of the motion semantics. Next, Fig.~\ref{fig:naturalness} shows the results of \textbf{Naturalness}. It is very clear that subjects regard on-manifold samples as more natural. This is understandable as manifold projection not only makes sure the poses are on the manifold, but also enforces the similarity of the dynamics between the attacked and original motion. Finally, the results of \textbf{Indistinguishability} are 89.90\% on average. BASAR even outperforms the white-box attack (80.83\%) in \cite{wang_smart_2020}. We further look into on-manifold vs off-manifold. Both samples are tested in Indistinguishability, but 94.63\% of the on-manifold samples fooled the users; while 84.69\% of the off-manifold samples fooled the users, showing that on-manifold samples are more deceitful. The off-manifold samples which successfully fool the users contain only very small deviations.

% 93.90\% on STGCN, 85.90\% on MSG3D

\subsection{Comparison}
Since BASAR is the very first black-box adversarial attack method on skeletal motions, there is no baseline for comparison. We therefore employ several methods that are closet to our approach. The first is SMART \cite{wang_smart_2020} which also attacks skeletal motions but is a white-box approach. Although it can deliver a black-box attack, it needs a surrogate model. According to their work, we choose HRNN \cite{Du:2015:HRN} and 2SAGCN \cite{Shi_2019_CVPR} as the surrogate models. The second baseline method is MTS \cite{harford2020adversarial} which is a black-box method but only on general multivariate time-series. It is the most similar method to BASAR but does not model the data manifold. Another baseline method is BA \cite{drendel_decisionaa_2018} which attacks image data. Since it is also a black-box approach and focuses on the boundary attack, we also include it in the comparison. We choose HDM05 and NTU for comparisons. For each comparison, we randomly select 1200 and 490 motions from NTU and HDM05. Since MTS is not designed for untargeted attack, we only compare BASAR with it on the targeted attack. 
%For SMART, we randomly select 1200 and 490 results on each attack on NTU and HDM05 dataset.\HW(how many motions?). In addition, since MTS is not designed for untargeted attack, we attack every motion with a specific label \HW(how did you choose this label?) in validation set \HW(again, how many motions are in this validation set?) of NTU and HDM05. To compare with BA, we attack the same target motions (\HW(don't get it. Why not attack the same target motions using SMART and MTS?)) with BASAR.

Tab.~\ref{tab:comparison1} lists the success rates of all methods. BASAR performs the best and often by big margins. In the targeted attack, the highest among the baseline methods is merely 30.31\% which is achieved by SMART on HDM05/MSG3D. BASAR achieves 100\%. In the untargeted attack, the baseline methods achieve higher performances but still worse than BASAR. SMART achieves as high as 99.33\% on NTU/MSG3D. However, its performance is not reliable as it highly depends on the chosen surrogate model, which is consistent with \cite{wang_smart_2020}. In addition, we further look into the results and find that SMART's results are inconsistent. When the attack is transferred, the result labels are often different from the labels obtained during the attack. 

We find that BA can also achieve 100\% success. However, BA is designed to attack image data and does not consider the data manifold. We therefore compare detailed metrics and show the results in Tab.~\ref{tab:comparison2}. BA is in general worse than BASAR under every metric. The worst is the bone-length constraint violation. Visually, the skeletal structure cannot be observed at all. This happens for both the untargeted and the targeted attack across all datasets and models. This is understandable because BA's assumption is that perturbations in all directions have the same effect, while BASAR assumes that in-manifold perturbations provide better visual quality.

% Also, SMART can only attack motions into classes with very similar semantics, e.g. take off a hat to take off a cap. Finally, there is a lack of diversity in the attack. Although the attack is untargeted, most adversarial motion are attacked to the same label, like HRNN/NTU/Untargeted Attack/MSG3D, 99.5\% results is attacked as 'touch other person's pocket'. In 2sgcn/ntu/untargeted/sgn, 52.3\% results are attack as 'walking towards each other'. I'm not sure whether these details would be helpful to the evaluation
%Comparsion results
\begin{table}[]
\centering
\setlength{\tabcolsep}{0.7mm}{
\begin{tabular}{|c|c|c|c|}
\hline
Targeted   & Attack Method  & HDM05    & NTU      \\  \hline
\multicolumn{1}{|c|}{\multirow{4}{*}{STGCN}}                                    & BASAR                          & 100.00\% & 100.00\% \\  
\multicolumn{1}{|c|}{}                                                           & MTS                    & 3.27\%   & 12.00\%  \\  
\multicolumn{1}{|c|}{}                                                           & SMART(HRNN)                    & 3.20\%   & 2.27\%   \\  
\multicolumn{1}{|c|}{}                                                           & SMART(2SAGCN)                   & 2.33\%   & 0.19\%   \\ \hline
\multirow{4}{*}{MSG3D}                                                          & BASAR                          & 100.00\% & 100.00\% \\  
                                                                                 & MTS                    & 2.18\%   & 12.90\%  \\  
                                                                                 & SMART(HRNN)                    & 30.31\%  & 1.16\%   \\  
                                                                                 & SMART(2SAGCN)                   & 2.46\%   & 0.00\%   \\ \hline
\multirow{4}{*}{SGN}                                                             & BASAR                          & 100.00\% & 100.00\% \\  
                                                                                 & MTS                    & 2.91\%   & 0.00\%   \\  
                                                                                 & SMART(HRNN)                    & 29.69\%  & 1.42\%   \\  
                                                                                 & SMART(2SAGCN)                   & 3.28\%   & 1.83\%   \\ \hline\hline
Untargeted   & Attack Method  & HDM05    & NTU      \\  \hline
\multirow{3}{*}{STGCN}                                                          & BASAR                          & 100.00\% & 100.00\% \\  
                                                                                 & SMART(HRNN)                    & 66.87\%  & 89.25\%  \\  
                                                                                 & SMART(2SAGCN)                   & 86.10\%  & 12.88\%  \\ \hline
\multirow{3}{*}{MSG3D}                                                          & BASAR                          & 100.00\% & 100.00\% \\  
                                                                                 & SMART(HRNN)                    & 86.88\%  & 99.33\%  \\  
                                                                                 & SMART(2SAGCN)                   & 88.73\%  & 3.08\%   \\ \hline
\multirow{3}{*}{SGN}                                                             & BASAR                          & 100.00\% & 100.00\% \\  
                                                                                 & SMART(HRNN)                    & 89.25\%  & 98.25\%  \\  
                                                                                 & SMART(2SAGCN)                   & 0.41\%   & 97.75\%  \\ \hline
\end{tabular}}
\caption{Attack success rate comparison with baseline methods.}
\label{tab:comparison1}
\end{table}
% Boundary Attack results
\begin{table}[]
\centering
\setlength{\tabcolsep}{1mm}{
\begin{tabular}{|c|c|ccccc|}
\hline
Models&  & l$\uparrow$ & $\Delta$a$\uparrow$ & $\Delta\alpha\uparrow$ & $\Delta$B/B$\uparrow$ & OM$\downarrow$ \\
\hline
\multirow{2}{*}{STGCN} & UA & 1.44 & 0.65& 4.74 & 10.60\% & 0.00\% \\
& TA & 8.83 & 0.17 & 1.60 & 8.56\% & 0.00\%\\
\hline
\multirow{2}{*}{MSG3D} & UA & 1.17  & 0.36 & 2.81  & 6.00\% & 0.00\% \\
& TA & 7.93  & 0.10 & 1.07  & 7.49\% & 0.00\%\\
\hline
\multirow{2}{*}{SGN} & UA & 13.35 & 3.45 & 21.96 & 75.11\% & 0.00\% \\
& TA & 15.40 & 1.52 & 11.77 & 29.45\% & 0.00\%  \\  
\hline\hline
\multirow{2}{*}{STGCN} & UA & 1.04 & 0.47 & 1.97 & 235.10\% & 0.00\%  \\
& TA & 0.41 & 0.08 & 0.37 & 35.00\% & 0.00\%  \\ 
\hline
\multirow{2}{*}{MSG3D} & UA & 1.24 & 1.73 & 2.38 & 911.7\% & 0.00\%  \\
& TA & 0.27 &0.07  & 0.34 & 25.72\% & 0.00\%  \\ 
\hline
\multirow{2}{*}{SGN} & UA & 0.22 & 0.28 & 1.26  & 125.57\% & 3.60\% \\
& TA & 0.42 & 0.15 & 0.66  & 65.31\% & 0.17\%  \\ 
\hline
\end{tabular}}
\caption{Boundary Attack (BA) on HDM05 (top), NTU (bottom). UA/TA refers Untargeted/Targeted Attack.} %All achieved 100\% successful rate. We show the the averaged joint position deviation ($l$), averaged joint acceleration deviation ($\Delta$a), averaged joint angular acceleration deviation ($\Delta\alpha$), averaged bone-length deviation percentage ($\Delta$B/B), on-manifold sample percentage (OM). UA/TA refers Untargeted/Targeted attack.}
\label{tab:comparison2}
\vspace{-0.3cm}
\end{table}

\section{Discussion and Conclusion}
To systematically investigate the vulnerability of state-of-the-art action recognizers, we proposed the very first black-box adversarial attack method which gives strong performance across datasets, models and attack modes. Quantitative evaluations demonstrate the high quality of the attack. Harsh perceptual studies show the imperceptibility of BASAR. Through comprehensive comparisons, BASAR outperforms existing methods by large margins in the success rate and quality. More broadly, we show, for the first time, the wide existence of on-manifold adversarial samples in skeletal motions. Based on the attack, we discuss what classifiers or features tend to be robust against attack. BASAR raises a serious yet unaddressed security concern that we expect will inform the future research. In future, we will investigate the perceptually-guided adversarial attack. We found that imperceptible perturbations form a small subspace. The interplay among the classification boundary, the data manifold and the imperceptible subspace is likely to lead to adversarial samples with higher quality. The interplay is also likely to shed light on effective defense strategies against imperceptible attack.

\noindent
\textbf{Acknowledgements:} This   project   has   received   funding   from   the   European
Union’s Horizon 2020 research and innovation programme
under grant agreement No 899739 CrowdDNA, EPSRC (EP/R031193/1), NSF China (No. 61772462, No. U1736217), RCUK grant CAMERA (EP/M023281/1, EP/T014865/1), China Scholarship Council (201907000157), and the 100 Talents Program of Zhejiang University.

{\small
\bibliographystyle{ieee_fullname}
\bibliography{references}
}

\end{document}